\documentclass[letterpaper, 10 pt, conference]{ieeeconf}  

\IEEEoverridecommandlockouts                              
\usepackage{cite}
\usepackage{amsmath,amssymb,amsfonts}
\usepackage{algorithmic}
\usepackage{graphicx}
\usepackage{textcomp}
\usepackage{xcolor}
\usepackage{multirow}
\usepackage{booktabs}
\usepackage{url}
\usepackage{amsmath}
\usepackage{mathrsfs}
\usepackage{makecell}
\usepackage{flushend}
\usepackage[section]{placeins}

\title{\LARGE \bf
Monocular Localization with Semantics Map for Autonomous Vehicles
}

\author{Jixiang Wan$^{1,2,\ast,\dagger}$, Xudong Zhang$^{1,\dagger}$, Shuzhou Dong$^{1}$, Yuwei Zhang$^{1}$,\\ Yuchen Yang$^{1}$, Ruoxi Wu$^{1}$, Ye Jiang$^{1}$, Jijunnan Li$^{1}$, Jinquan Lin$^{1}$, Ming Yang$^{2}$
\thanks{$^{1} $OPPO Research Institute, Shanghai, China.}
\thanks{$^{2} $Department of Automation, Shanghai Jiao Tong University, Shanghai, China.}
\thanks{${\dagger}$ Authors contributed equally to this work.}
\thanks{${\ast}$ indicates corresponding author. Contact: wanjixiang@oppo.com}
}

\begin{document}

\maketitle
\thispagestyle{empty}
\pagestyle{empty}

\begin{abstract}

    Accurate and robust localization remains a significant challenge for autonomous vehicles. The cost of sensors and limitations in local computational efficiency make it difficult to scale to large commercial applications. Traditional vision-based approaches focus on texture features that are susceptible to changes in lighting, season, perspective, and appearance. Additionally, the large storage size of maps with descriptors and complex optimization processes hinder system performance. To balance efficiency and accuracy, we propose a novel lightweight visual semantic localization algorithm that employs stable semantic features instead of low-level texture features. First, semantic maps are constructed offline by detecting semantic objects, such as ground markers, lane lines, and poles, using cameras or LiDAR sensors. Then, online visual localization is performed through data association of semantic features and map objects. We evaluated our proposed localization framework in the publicly available KAIST Urban dataset and in scenarios recorded by ourselves. The experimental results demonstrate that our method is a reliable and practical localization solution in various autonomous driving localization tasks.

\end{abstract}

\section{INTRODUCTION}
In recent times, autonomous vehicles have received increasing attention from both academia and industry. Accurate and robust self-localization is critical for autonomous driving and serves as the foundation for subsequent applications, which include path planning, cooperative driving, map updating, and more. Although centimeter-level localization accuracy is now achievable in many scenarios by utilizing high-precision sensors like GPS-RTK and LiDAR, their expensive hardware costs create obstacles for their widespread utilization. In contrast, vision sensors such as cameras with their mature processes and low expense are gaining significant attention in the realm of commercial autonomous driving solutions.

To achieve visual global localization, one popular approach is to solve the PnP problem which creates associations between the 2D features tracked in the current image and the 3D features in the pre-constructed Structure From Motion (SFM) map. To ensure success under varying viewpoints and lighting, it is crucial that the extracted visual features are highly repeatable and consistent.
Recent studies like \cite{sarlin2019coarse,li2020dxslam} have implemented learnable descriptors and matching strategies based on deep learning for good performance. Other researchers such as in \cite{zhou2020da4ad, sarlin2021back}
proposed end-to-end pose regression models, achieving outstanding results in their experiments. However, the generalization capability of these methods to new environments has not been demonstrated. Furthermore, the generating of complex descriptors can significantly increase map memory usage, which affects the computational efficacy, particularly in city-scale localization tasks.

The integration of semantic information has been shown to significantly enhance the accuracy and robustness of location estimation. Recent studies \cite{yang2019cubeslam,bescos2021dynaslam,liang2022semloc} have demonstrated the benefits of utilizing semantic information in the environment to improve the representation of visual features and simplify the computational requirements. Especially in autonomous driving scenarios, lightweight localization can be achieved by detecting semantic objects such as ground markers, lane lines, crosswalks and pole-like objects \cite{jeong2017road,li2020vision,qin2021light}. Compared with traditional visual features, these semantic features are widely available on urban roads and have long-term stability and robustness in the face of weather changes, light fluctuations, perspective changes, and occlusions caused by dynamic obstacles \cite{lu2017monocular,wang2021visual}. Furthermore, producing a semantic map using semantic objects instead of dense points can further reduce the cost of map distribution and storage.

Associating semantic cues from current observations and elements in a semantic map offers a promising solution for monocular visual localization in autonomous vehicles. However, there are several challenges to consider. On the one hand, standard vector High-Definition (HD) maps usually require specialized data-acquisition equipment and significant manpower for labeling. On the other hand, correctly transforming targets in 2D images to 3D real shapes presents a challenging problem due to dimension degradation defects. Therefore, this paper proposes a lightweight visual localization pipeline for autonomous driving, consisting of a semantic map constructor without manual annotation and a localization module using low-cost cameras and IMU devices.
The main contributions of this paper are summarized as follows:

\begin{itemize}
    \item We propose an enhanced inverse perspective mapping model that considers the rotation of camera, allowing for the accurate computation of bird’s-eye view images during motion.
    \item We propose an algorithm that facilitates the construction of global semantic maps using conventional LiDAR with minimal annotation assistance or supervision.
    \item We present a monocular localization algorithm based on common road visual semantic features and validate its effectiveness in real traffic scenarios.
\end{itemize}

\section{RELATED WORKS}

\subsection{Visual Localization}
VINS \cite{qin2018vins} and ORB-SLAM \cite{mur2015orb, mur2017orb, campos2021orb} are commonly used visual SLAM frameworks to achieve accurate trajectory measurements, integrating modules with feature point extraction and matching, keyframe bundle adjustment, loop closure detection, and map registration. Hloc \cite{sarlin2019coarse} constructs a global visual localization framework that includes image retrieval, local feature matching, and pose regression. Nonetheless, The real-time localization at city-scale presents a challenge for them.

LaneLoc \cite{schreiber2013laneloc} is one of the pioneers that utilizes lane lines in combination with a prior semantic maps. TM3Loc \cite{wen2022tm} propose a tightly-coupled vehicle localization framework using semantic landmark matching in a HD Map. RSCM \cite{kim2022road} attempts to resolve the underdetermined problem in registration methods by dividing lane segments into shapes and curves. Dt-loc \cite{zhang2021dt} proposes distance transforms of the semantic detection to enable the differentiable data association process to achieve high localization precision. LAVIL \cite{li2022lidar} explores the limit of visual semantic localization with the aid of LiDAR odometry.
Improving the accuracy and reducing the cost of manual production of the prefabricated maps are effective ways to advance semantic localization approaches.

\subsection{LiDAR SALM}
LiDAR has the ability to detect the real scale and location information of objects, which can significantly enhance the creation of high-precision semantic maps. Most existing LiDAR SLAM works can be traced back to the LOAM algorithm \cite{zhang2014loam}, which proposes approaches for extracting valid feature points and registering a global map. The subsequent Lego-LOAM \cite{shan2018lego} method disregards ground points to expedite computation and incorporates a loop closure detection module to reduce the long-term drift. LIOM \cite{ye2019tightly} introduces IMU pre-integration into odometry and proposes a LiDAR-imu tightly coupled SLAM method. FAST-LIO \cite{xu2021fast} addresses the issue of motion distortion during point cloud scanning by utilizing IMU measurements to compensate for motion distortion, while improving the Kalman gain formula formulation to reduce the computational effort of iterative optimization. FAST-LIO2 \cite{xu2022fast} proposes an incremental k-d tree data structure, ikd-Tree, to improve the search efficiency. It makes large-scale dense point cloud computing possible.

\section{PROPOSED APPROACH}
\begin{figure}[htbp]
    \centering
    \includegraphics[width=0.45\textwidth]{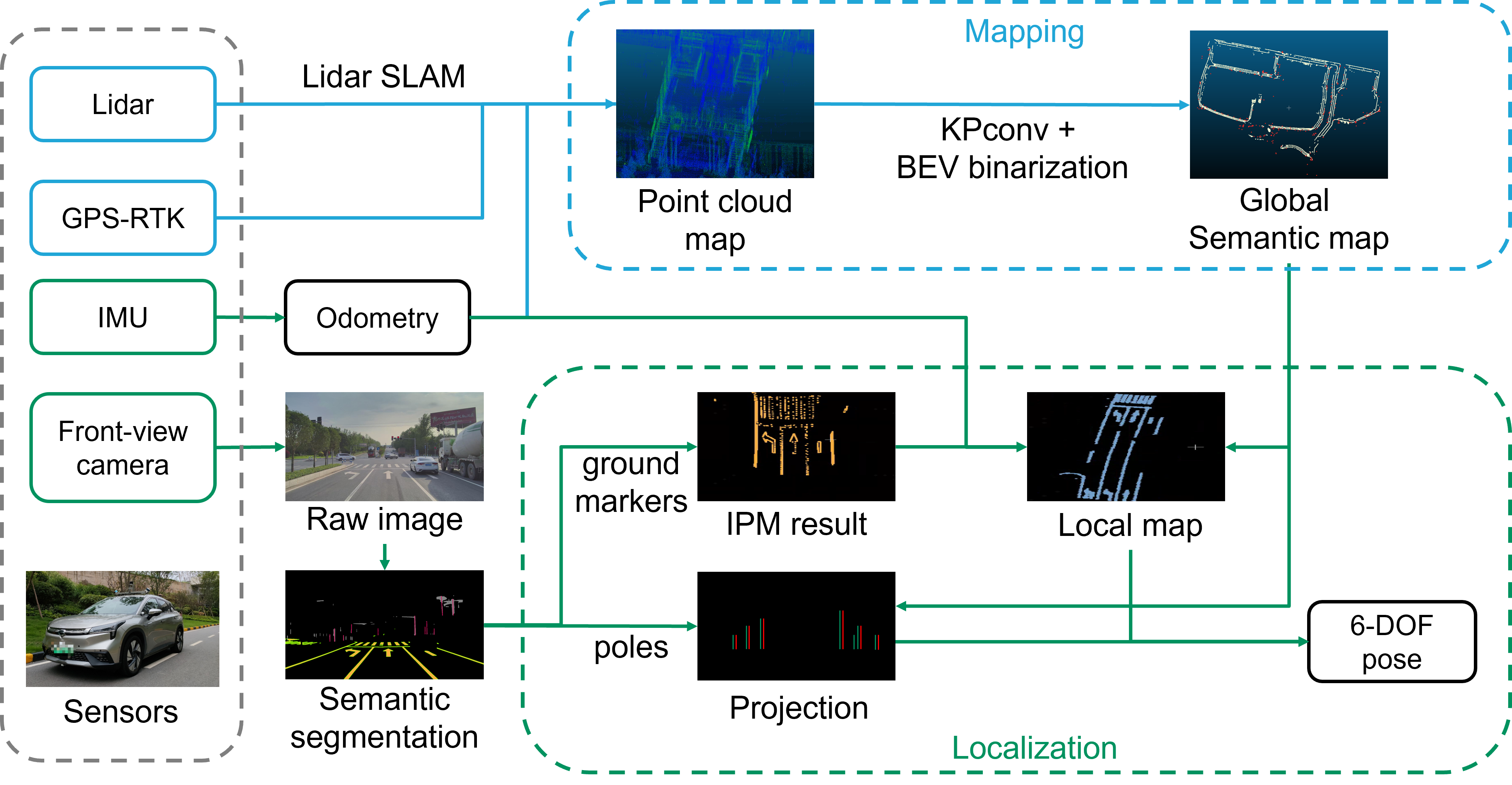}
    \caption{Illustration of the system structure.
        The upper part illustrates the construction of global semantic map, and the lower part is the vehicle localization process is through the monocular camera.
    }
    \label{fig:pipeline}
\end{figure}

In this work, we present a visual localization method based on semantic map, as shown in Fig. \ref{fig:pipeline}. The system consists two parts: (1) Global semantic map generation. The data collected from roads by vehicles equipped with LIDAR, GPS-RTK and IMU, or other navigation sensors, is utilized in creating point cloud maps using LiDAR SLAM.
The semantic features such as lane lines, lane signs, and pole-like objects are extracted from the point cloud to construct a semantic map. (2) Localization module. We use CNN to extract semantic information from the image captured by camera. Ground pixels (e.g. landmarks, crosswalks, lane lines) are used to construct a local map using inverse perspective mapping (IPM), and aligned with the global map. Pole-like objects (e.g. trunks, streetlights, poles of traffic lights and billboards) on the semantic map are projected onto the image to create line matching. The vehicle's 6-DOF pose can be obtained by minimizing the global reprojection error.

\subsection{Semantic Map}

\begin{figure}
    \includegraphics[width=0.45\textwidth]{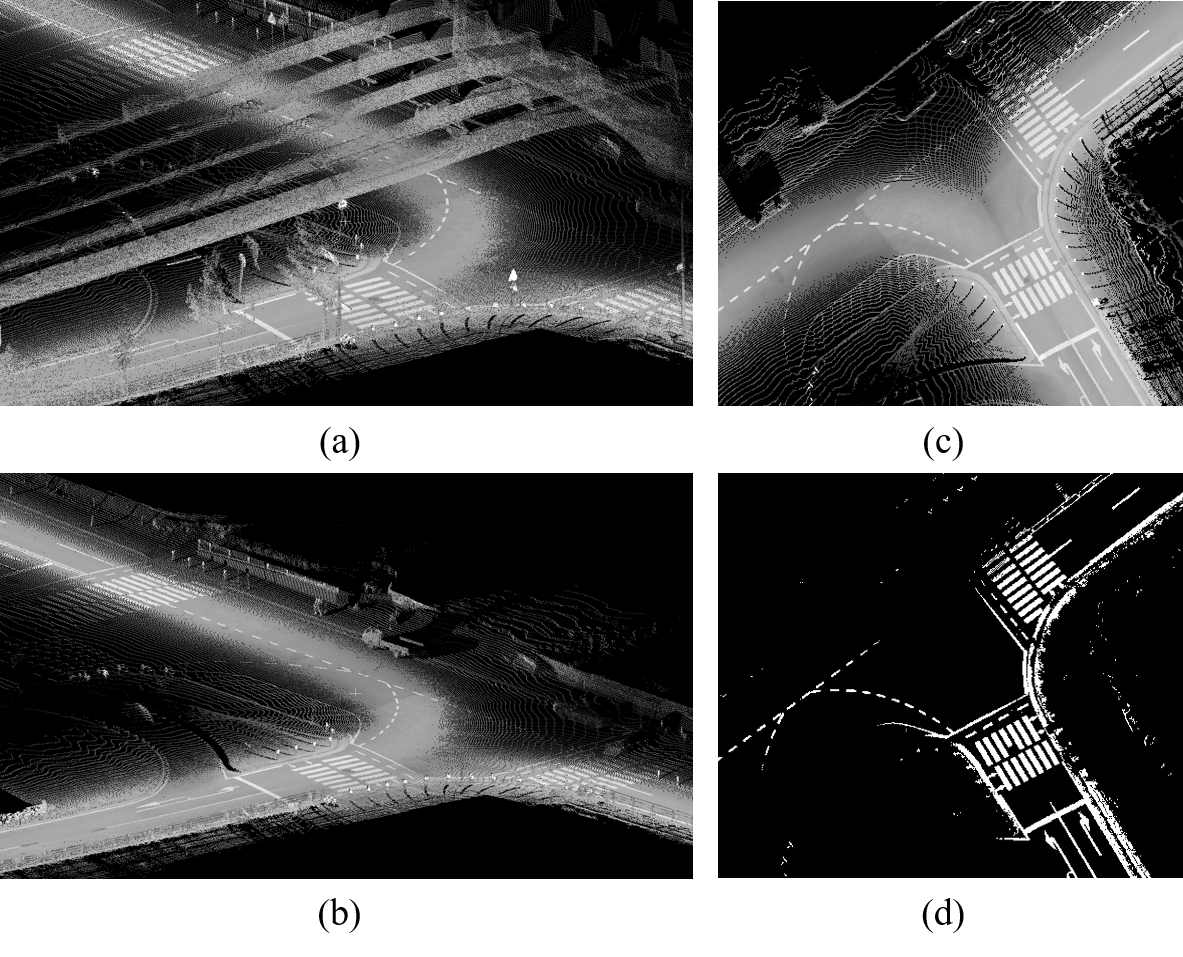}
    \centering
    \caption{Point cloud map generation and BEV segmentation. (a) shows the original point cloud map. (b) is the ground point cloud produced by LiDAR SLAM. (c) provides an example of BEV image, where each pixel corresponds to a 10 cm voxel. (d) displays the OTSU binarization results, which preserves high-contrast features on roads, including lane lines and markers.}
    \label{fig:bev}
\end{figure}

With the improved FAST-LIO2 algorithm \cite{xu2022fast} by fusing GPS-RTK information in the pose graph optimization module to ensure global location accuracy, the data collected by LiDAR is registered as a high-precision point cloud map. From which, we segment the pole-like semantic, and extract the two endpoints of each pole using Euclidean clustering and RANSAC linear fitting. Ground point clouds are extracted from pre-trained KPConv models \cite{thomas2019kpconv} and plane-growth method. To accurately segment the ground marks,
we project KPConv segmentation results onto the BEV plane, with the road surface point cloud's reflectivity treated as pixel values. Here, we employ the OTSU algorithm \cite{None1979A} to further binarize reflectivity values, enabling the isolation of clear lane markings and road surfaces. Finaly, We apply the mapping relationship between 3D point cloud and the BEV image to back-project the segmentation results into the 3D point cloud, enabling the 3D spatial semantic segmentation of the relevant elements, as depicted in Fig. \ref{fig:bev}.

\subsection{Image Segmentation}
\begin{figure}[htbp]
    \centering
    \includegraphics[width=0.45\textwidth]{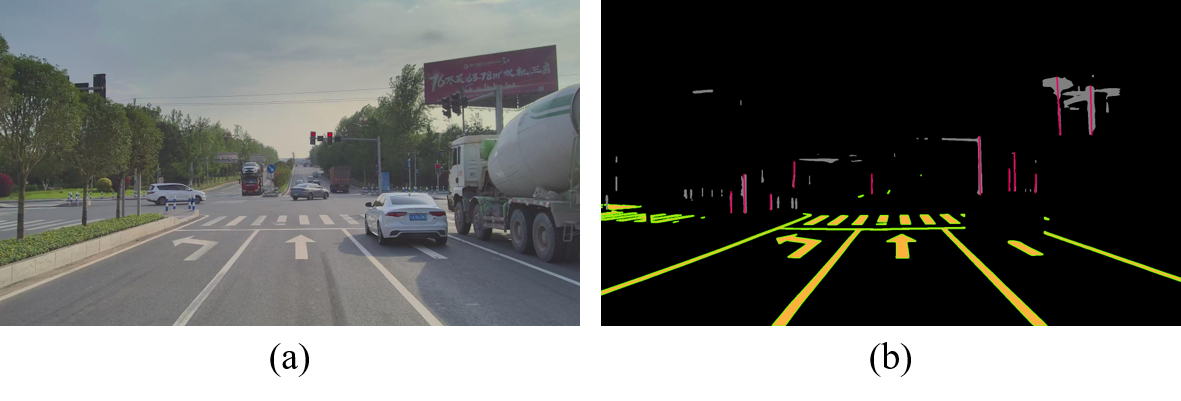}
    \caption{Image segmentation. (a) is the raw image captured by front-view camera. (b) is the semantic segmentation result. The orange and gray pixels indicate ground markers and poles, respectively. Green pixels highlight the outline of the ground markers and red pixels indicate the fitted straight lines of the poles. Note that short poles are discarded to avoid bringing in noise.}
    \label{fig:img_seg}
\end{figure}

The first step of localization is the semantic segmentation of images. We divide all the semantics into three categories: ground markers, poles, and background. A lightweight model, BiSeNetV2 \cite{yu2021bisenet}, is selected to segment the necessary semantic features. To improve the computational efficiency of pixel projection, OpenCV \cite{2000The} is used to extract all ground markings contours instead of using the entire semantic masks. This approach is favored because the position information of the contour can provide equivalent spatial constraints as whole marking pixels. Each pole instance is fitted as a straight line using the least squares method, which facilitates calculating the distance from the map point to the fitted pole. Fig. \ref{fig:img_seg} illustrates a visualization of image segmentation in real traffic scenes.

\subsection{Inverse Perspective Transformation}



\begin{figure}[htbp]
    \centering
    \includegraphics[width=0.45\textwidth]{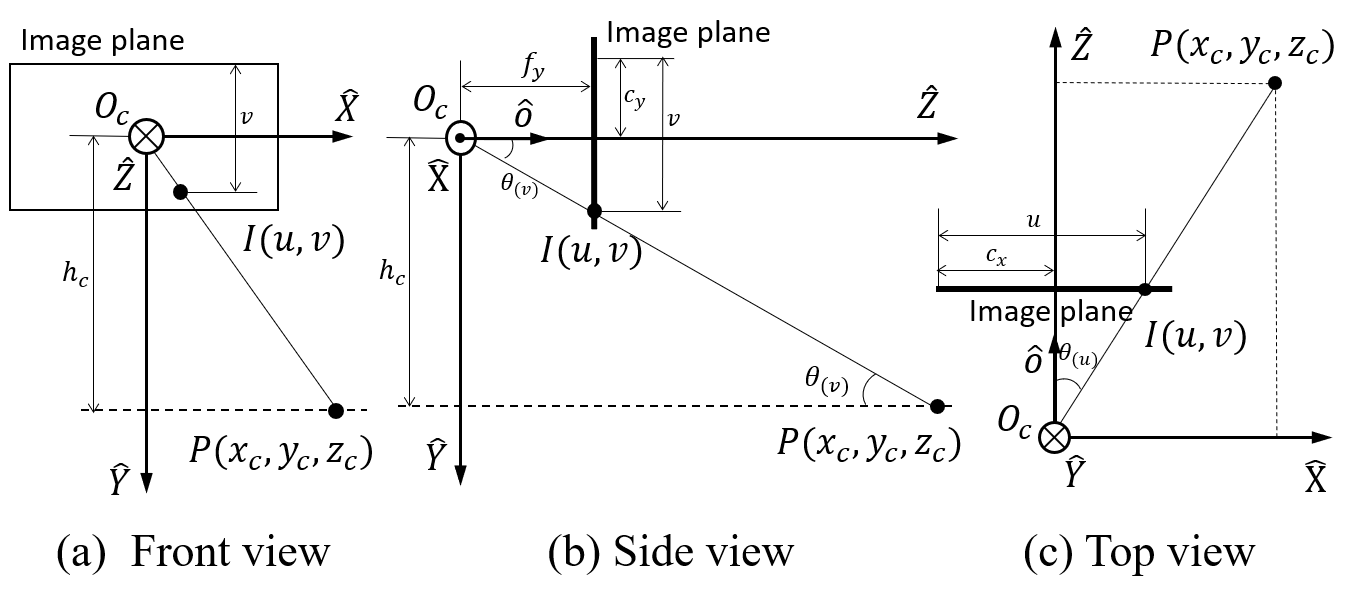}
    \caption{The schematic of basic IPM model.}
    \label{fig:IPM_model}
\end{figure}

After segmentation, the ground markers are transformed from the image plane to the vehicle coordinate system. This process can be executed through the IPM algorithm. Fig. \ref{fig:IPM_model} provides the conventional IPM model using physical parameters of the pinhole camera. The projection of point $P$ in the ground plane to point $I$ in the image plane is shown from three views.

Based on the principles of projection for the pinhole camera, the translation from the point $[x_c, y_c, z_c]^T \in R^3$ to the pixel $[u, v]^T \in I^2$ can be described as Eq. \ref{eq: ipm_projection}.

\begin{equation}
    Z\times\begin{bmatrix}u\\v\\1\end{bmatrix} = K\times\begin{bmatrix}x_c\\y_c\\z_c\end{bmatrix} = \begin{bmatrix}f_x&s&c_x\\0&f_y&c_y\\0&0&1\end{bmatrix}\times\begin{bmatrix}x_c\\y_c\\z_c\end{bmatrix}
    \label{eq: ipm_projection}
\end{equation}

\noindent where $K$ is the intrinsic matrix of camera, $f_x$ and $f_y$ represent the focal length, $s$ is the skewness factor, and $(c_x, c_y)$ denotes the optical center of camera.



It is a reasonable assumption that a vehicle's wheels remain in contact with the ground while it is being driven on the road. This implies that the vertical height $h$ between the ground and the optical center of the camera mounted on the vehicle remains constant. As shown in Fig. \ref{fig:IPM_model}(b), the tilt angle between line $\overline{O_cP}$ and optical axis $\hat{o}$ is determined by the vertical coordinate $v$ of point $I$. This angle is represented as $\theta(v)$, and the geometric relation can be expressed via the equation $\tan(\theta(v)) = \dfrac{z_c}{y_c}$. From Eq. \ref{eq: ipm_projection}, $\theta(v)$ can be derived as:

\begin{equation}
    \theta(v) = \arctan(\dfrac{f_y}{v-c_y})
    \label{eq: ipm_theta_v}
\end{equation}

Similarly, from the geometric relations illustrated in \ref{fig:IPM_model}(c), we can deduce $\theta(u)$.

\begin{equation}
    \begin{aligned}
        \theta(u) & = \arctan(\dfrac{x_c}{z_c}) = \arctan(\dfrac{f_y (u-c_x)- s (v-c_y)}{f_x f_y})
    \end{aligned}
    \label{eq: ipm_theta_u}
\end{equation}

Eq. \ref{eq: ipm_theta_v} and \ref{eq: ipm_theta_u} define the fixed mapping relationship between the position of the point $P=[x_c, y_c, z_c]^T$ and $I=[u, v]^T$.

\begin{figure}[htbp]
    \centering
    \includegraphics[width=0.45\textwidth]{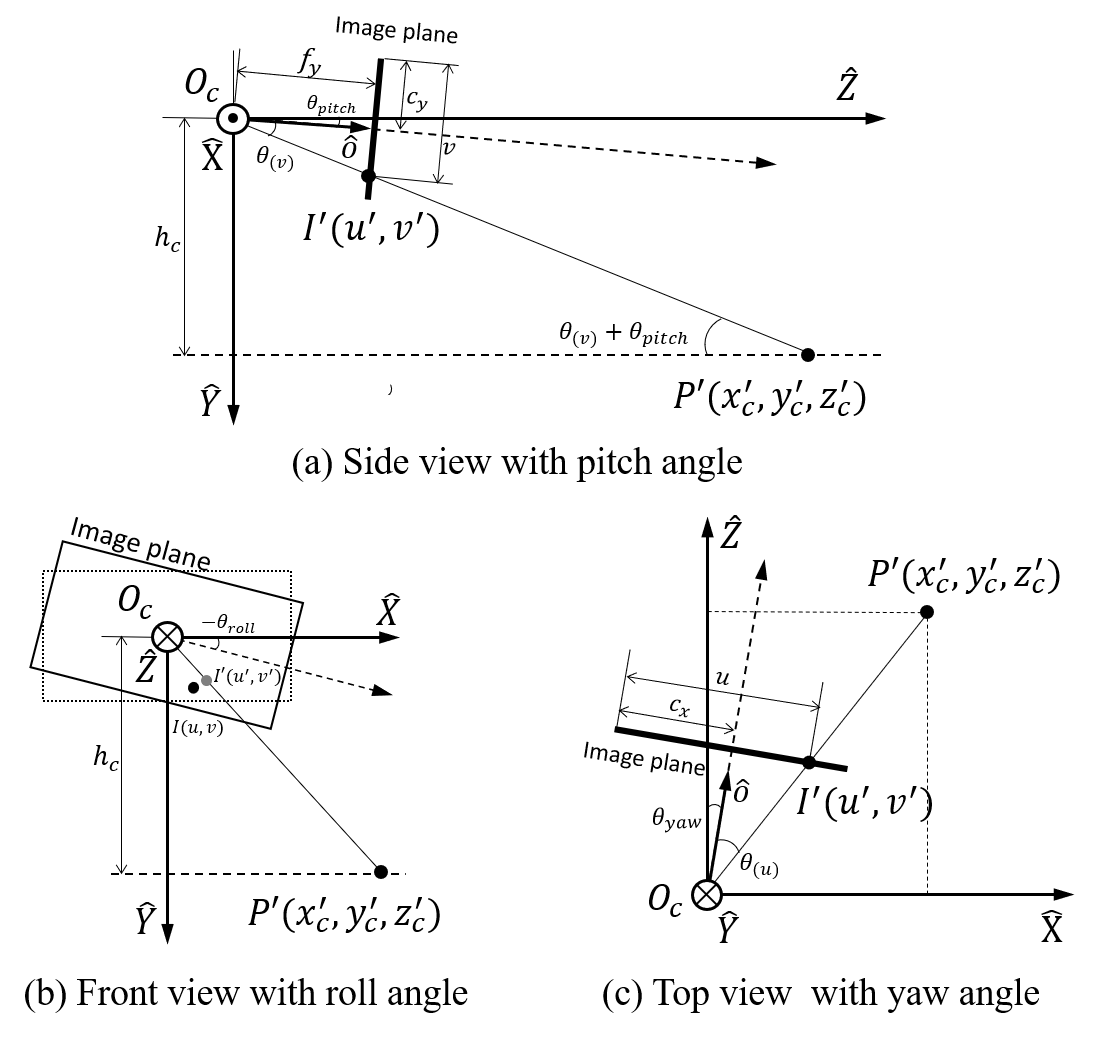}
    \caption{The schematic of the enhanced IPM model with roll, pith, and yaw angles compensation.}
    \label{fig:IPM_model_with_rpy}
\end{figure}

However, this basic IPM model is limited to the ideal situation where the ground surface is perfectly horizontal and the camera optical axis $\hat{o}$ is strictly parallel to the ground. In reality, vehicle motion induces camera rotation along the $\hat{X}$, $\hat{Y}$ and $\hat{Z}$ axes, denoted as $\theta_{roll}$, $\theta_{pitch}$ and $\theta_{yaw}$. The enhanced IPM model with rotation angle compensation is shown in Fig. \ref{fig:IPM_model_with_rpy}. The projection position $P$ in the ground plane shifts to $P^{'}=[x_{c}^{'}, y_{c}^{'}, z_{c}^{'}]^T$.

The effect of $\theta_{roll}$ can be visualized as rotating the image plane along the optical axis $\hat{o}$, as shown in Fig. \ref{fig:IPM_model_with_rpy}(b).
Therefore, the equivalent image mapped point can be obtained by rotating $I^{'}=[u^{'}, v^{'}]^T$ with an angle equal to $-\theta_{roll}$.
The transformation from $I$ to $I^{'}$ can be described as follows.

\begin{equation}
    \begin{bmatrix}u^{'}\\v^{'}\end{bmatrix} = \begin{bmatrix}\cos(\theta_{roll})&\sin(\theta_{roll})\\-\sin(\theta_{roll})&\cos(\theta_{roll})\end{bmatrix}\times\begin{bmatrix}u\\v\end{bmatrix}
    \label{eq: ipm_uv_}
\end{equation}

As shown in the side view from Fig. \ref{fig:IPM_model_with_rpy}(a), $\theta_{pitch}$ causes
an inclination with axis $\hat{Z}$ of camera coordinate system.
This expressions of $z_{c}^{'}$ is deduced in Eq. \ref{eq: ipm_z_}.

\begin{equation}
    \begin{aligned}
        z_{c}^{'} & = y_{c}^{'}\cdot \cot(\theta(v^{'})+\theta_{pitch})                                                   \\
                  & = h\cdot \dfrac{1-\tan(\theta(v^{'})) \tan(\theta_{pitch})}{\tan(\theta(v^{'}))+\tan(\theta_{pitch})} \\
    \end{aligned}
    \label{eq: ipm_z_}
\end{equation}

Note that $z_{c}^{'}$ depends only on the variables $v^{'}$ of the pixel point $I^{'}=[u^{'}, v^{'}]^T$ and $\theta_{pitch}$. We derive $x_{c}^{'}$ using a proportional expression between $x_{c}^{'}$ and $z_{c}^{'}$ as illustrated in Fig. \ref{fig:IPM_model_with_rpy}(c).

\begin{equation}
    \begin{aligned}
        x_{c}^{'} & = z_{c}^{'}\cdot \tan(\theta(u^{'})+\theta_{yaw})                                                         \\
                  & = z_{c}^{'}\cdot \dfrac{\tan(\theta(u^{'}))+\tan(\theta_{yaw})}{1-\tan(\theta(u^{'})) \tan(\theta_{yaw})} \\
    \end{aligned}
    \label{eq: ipm_x_}
\end{equation}

Furthermore, the installation of the camera results in an initial deviations $\theta_{roll,0}$, $\theta_{pitch,0}$, and $\theta_{yaw,0}$. These deviations are fixed and can be obtained through factory calibration. As a result, the compensation equations of the enhanced IPM can be derived from Eq. \ref{eq: ipm_uv_}, \ref{eq: ipm_z_} and \ref{eq: ipm_x_}.

\begin{equation}
    \begin{cases}
        u^{'} = u\cdot \cos(\theta_{roll,0} +\theta_{roll}) + v\cdot \sin(\theta_{roll,0} +\theta_{roll})                                                       \\
        v^{'} = -u\cdot \sin(\theta_{roll,0} +\theta_{roll}) + v\cdot \cos(\theta_{roll,0} +\theta_{roll})                                                      \\
        \tan(\theta(u^{'})) = \dfrac{f_y (u^{'}-c_x)- s (v^{'}-c_y)}{f_x\cdot f_y}                                                                              \\
        z_{c}^{'} = h\cdot \dfrac{1-\tan(\theta_{pitch,0}+\theta_{pitch}) \dfrac{f_y}{v^{'}-c_y}}{\tan(\theta_{pitch,0}+\theta_{pitch})+\dfrac{f_y}{v^{'}-c_y}} \\
        x_{c}^{'} = z_{c}^{'}\cdot \dfrac{\tan(\theta_{yaw,0} + \theta_{yaw})+\tan(\theta(u^{'}))}{1-\tan(\theta_{yaw,0} + \theta_{yaw}) \tan(\theta(u^{'}))}   \\
        y_{c}^{'} = h
    \end{cases}
    \label{eq: ipm_uvxyz_}
\end{equation}

\begin{figure}[htbp]
    \centering
    \includegraphics[width=0.45\textwidth]{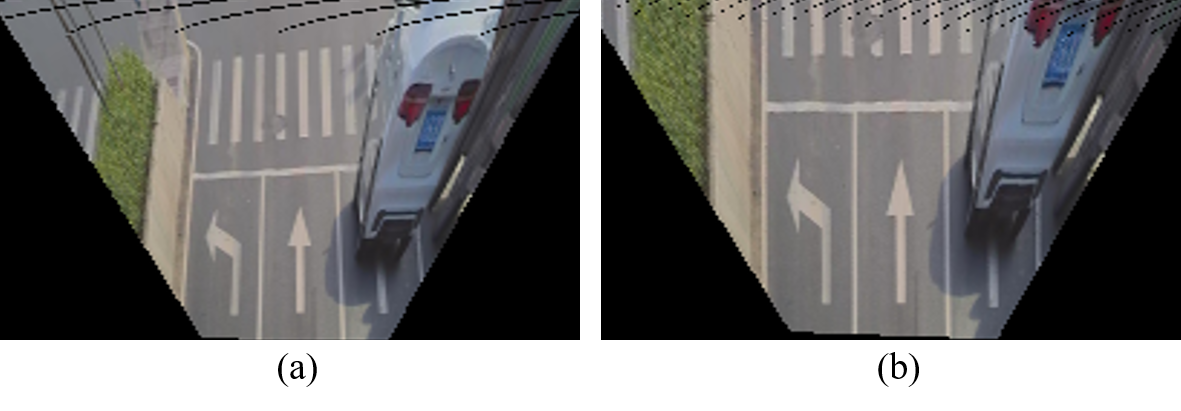}
    \caption{(a) is the BEV image transformed by the vanilla IPM. (b) is the enhanced IPM result with deflection angle compensation $roll,pitch,yaw=(0.8^{\circ},-1.9^{\circ},-1.2^{\circ})$.}
    \label{fig:img_ipm}
\end{figure}

In the real-world driving scenario, The deflection angles $(\theta_{roll}, \theta_{pitch}, \theta_{yaw})$ of the moving vehicle are computed via the integration of IMU data. Subsequently, the IPM model with rotation compensation is used to compute the projected coordinates of specific pixels and accurately restore the their 3D position in space. Fig. \ref{fig:img_ipm}(a) shows the distorted BEV image of the vanilla IPM model. On the other hand, Fig. \ref{fig:img_ipm}(b) presents the result of the enhanced IPM model with angle compensation. This illustrates the substantial distortion in the BEV image from even considerably small variations in angle during motion.

\subsection{Optimization Solver}
Before optimizing pose from the $k$-th image frame, the vehicle state must be prepared, including the prior pose as well as the position of the ground markers and pole features. The iterative nonlinear optimization method is then used to match the current features with the global semantic map, resulting in the current pose of the vehicle.

\textbf{Prior Pose.}
Vins-mono \cite{qin2018vins} propose an visual inertial odometry (VIO) method, which offers the vehicle's relative position and rotation. To improve the accuracy of the prior pose, the relative pose transformation between frame $k$ and $k-1$ is computed, and integrated to the semantic localization result of the previous frame. This helps to minimize the cumulative error caused by IMU integration. The prior pose for $k$-th frame, denoted as $T_{k}^{*}$, is expressed in the following formula.

\begin{equation}
    T_{k}^{*} = T_{k-1} (\hat{T}_{k-1})^{-1} \hat{T}_{k}
    \label{eq: T_init}
\end{equation}

\noindent where $T_{k-1}$ indicates the localization result of the previous frame obtained by our semantic localization algorithm. $\hat{T}_{k-1}$ and $\hat{T}_{k}$ are the VIO results for the corresponding frames respectively.

\textbf{Ground Markers Representation.}
In the $k$-th image frame, we preserve the contoured pixels of ground markers. we designate the positions of
$m$ points in lane marking contours as $P_{lane,k}^{I}=\{\langle p_i, lane \rangle \}_{i=1}^{m}$, where $p_i=[u_i,v_i]^T$ is pixel coordinate.
Therefore, the lane marking points in the vehicle coordinate system $\{\mathscr{V}\}$ can be represented as:

\begin{equation}
    P_{lane,k}^{V}=T_{C}^{V} \mathscr{M}_{ipm}(P_{lane,k}^{I})
    \label{eq: P_lane_V}
\end{equation}

\noindent where The matrix $T_{C}^{V}$ is the external parameters from $\{\mathscr{C}\}$ to $\{\mathscr{V}\}$ and remains constant. $\mathscr{M}_{ipm}(\cdot)$ represents the IPM model.

Due to the limited field of view and segmentation noise in a single image, we accumulate several frames of lane data by employing a sliding window. We generate a local semantic map that composed of the ground features from the most recent $c$ frames, while limiting its size to less than 50 meters. Subsequently, the local map can be transformed to the world coordinate system $\{\mathscr{W}\}$ with the prior pose $T_{V,k}^{W,*}$. We search the nearby points $\bar{P}_{lane}^{W}$ by building a KD-tree of the global semantic map, as formulated in Eq. \ref{eq: P_lane_W}.

\begin{equation}
    \bar{P}_{lane}^{W} \simeq P_{lane,k}^{W} = T_{V,k}^{W,*} \sum_{i=0}^{c}[T_{C}^{V} \mathscr{M}_{ipm}(p_{lane,k-i}^{I})]
    \label{eq: P_lane_W}
\end{equation}

Finally, we will only consider nearby points whose distance is less than a certain threshold (e.g. 1m). The loss is computed as follows:

\begin{equation}
    \mathscr{L}_{lane} \!=\! \!\sum\! \|P_{lane,k}^{W}-\bar{P}_{lane}^{W}\|^2 \!+\! \!\sum\! D(P_{lane,k}^{W}, \bar{L}_{lane}^{W})
    \label{eq: loss_lane}
\end{equation}

\noindent where $\bar{L}_{lane}^{W}$ denotes the fitted line using the 5 nearest points in the semantic map, and $D(\cdot)$ is used to measure the distance from a point to the line.

\textbf{Pole-like Objects Representation.}
When ground markings are not visible, relying solely on parallel lane lines fails to provide effective restraint in the forward direction of the vehicle. Pole-like objects (e.g. poles, lamp posts, tree trunks, etc.) are straight and perpendicular to the ground, which can be utilized to address this issue.

We use pairs of endpoints to denote $n$ poles in the semantic map as $\bar{P}_{pole}^{W} = \{\langle p_{1i},p_{2i}\rangle\}_{i=1}^{n}$, where each pole ${i}$ is represented by two endpoints $p_{1i}=[x_i,y_i,z_{1i}]^T$ and $p_{2i}=[x_i,y_i,z_{2i}]^T$. Furthermore, the poles are projected into $k$-th image frame as $\bar{P}_{pole}^{I}$ with prior pose and projection function.

\begin{equation}
    \bar{P}_{pole,i}^{I} = \frac{1}{z_i^C}K(T_{C}^{V})^{-1}(T_{V,k}^{W,*})^{-1} \bar{P}_{pole,i}^{W}
    \label{eq: P_pole_I_}
\end{equation}

\noindent where $z_i^C$ is the z-coordinate of point $i$ of the poles in camera coordinate $\{\mathscr{C}\}$.

For each endpoint projected onto the image, we find the closest straight line fitted by the segmentation result of pole-like objects. The distance from the endpoints $\bar{P}_{pole}^{I}$ to the corresponding fitted lines $L_{pole,i}^{I}$ is calculated as the residual.

\begin{equation}
    \mathscr{L}_{pole} = \sum_{i=0}^{n} [D(\bar{P}_{pole, i}^{I}, L_{pole,i}^{I})]
    \label{eq: loss_pole}
\end{equation}

Finally, the optimal global consistency matching is a nonlinear least squares problem, and the Ceres-Slover \cite{agarwal2012ceres} with the Levenberg-Marquardt (LM) algorithm is employed to solve the pose of the vehicle.

\begin{equation}
    T_{V,k}^{W} = \underset{T_{V,k}^{W,*}}{\arg \min} (\mathscr{L}_{lane} + \mathscr{L}_{pole})
    \label{eq: opt}
\end{equation}

\section{EXPERIMENTAL EVALUATION}

\subsection{Datasets}
The public KAIST dataset \cite{jeong2019complex} provides a variety of sensor data acquired from complex urban environments. We select some typical scenes of autonomous driving (i.e. suburban, urban, and highway) from sequences 26, 38, and 39. Among them, the given point cloud data from LiDAR is used to construct a global semantic map, while the left camera and IMU measurements are used for localization.

\begin{figure}[htbp]
    \centering
    \includegraphics[width=0.45\textwidth]{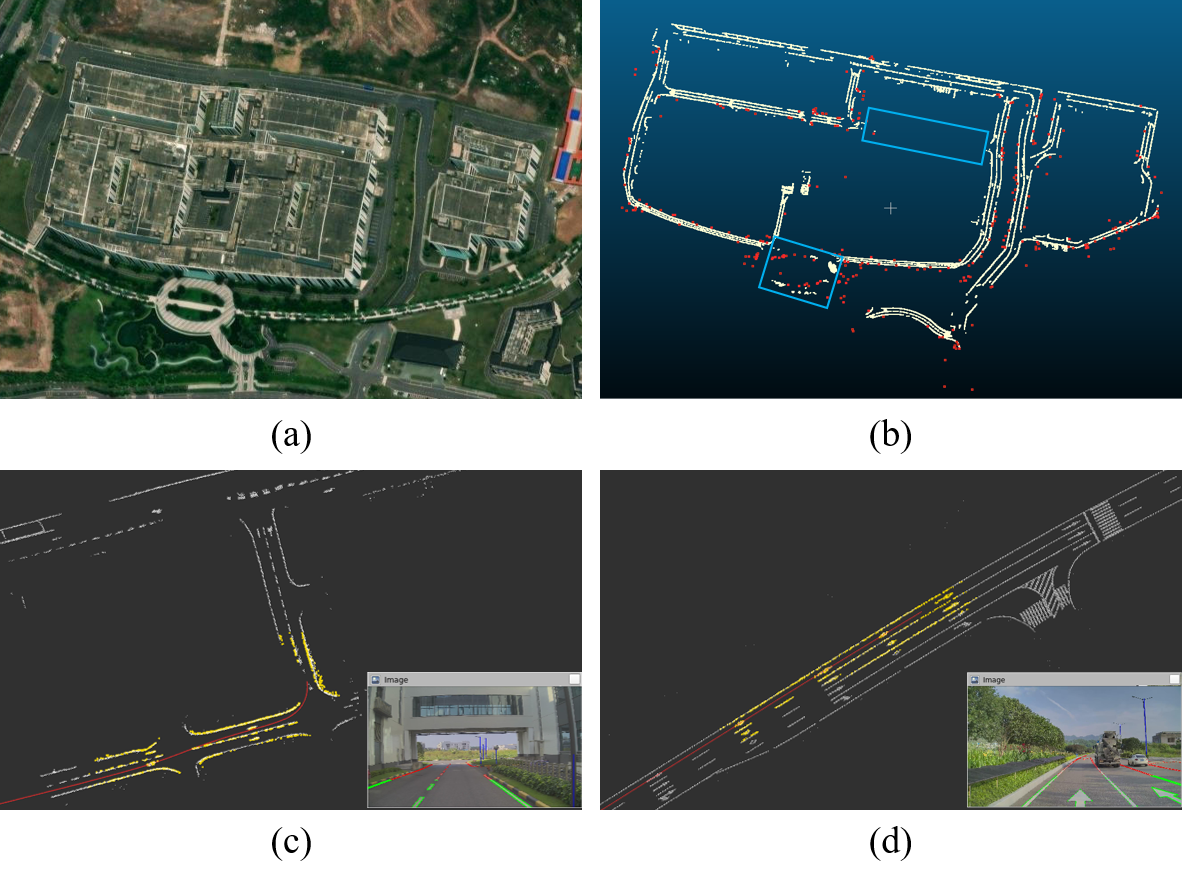}
    \caption{An illumination of the qualitative results. (a) shown the satellite map of industrial park area of our self-recorded dataset. (b) is the global semantic map of industrial park. The ground markers are drawn in yellow and the endpoints of poles are drawn in red. The blue boxes indicating the areas without sufficient lane semantic information. (c) is a visual example of real-time pose optimization of in the scenes of industrial park. (d) public roads. In (c) and (d), the white points denote the lane marking map dynamically loaded with grid zones, and the yellow points indicate the current local lane markers map, which is projected to world coordinate frame by optimized pose. The green pixels in images indicate the lane marking feature used in the current frame during the pose estimation. Due to perspective noise, the pixels that are too far from the camera are discarded during the optimization process and marked as red. The blue lines represent the fitted pole-like features.}
    \label{fig:illus_loc}
\end{figure}

In addition, we record a dataset covering a entire industrial park and several surrounding public roads, which add up to an approximately 6km-long road network in Chongqing, China. Fig. \ref{fig:illus_loc}(a) shows the satellite map of the selected area. The dataset is collected by our self-driving cars equipped with a front-view camera, LiDAR, GPS-RTK, and IMU. We use LiDAR data to construct the point cloud map, and treat the GPS-RTK as the ground truth of localization.

\subsection{Visual localization accuracy}
To evaluate the performance of our system, we compared it against other semantic localization algorithms, including CL+PA \cite{liao2020coarse}, PC semantic \cite{stenborg2018long}, and fusion SFM \cite{li2020vision} on the KAIST dataset. Following the benchmark, we consider the localization accuracy on the x, y directions, as well as the heading (yaw) angle. We used the root mean squared error of absolute trajectory error (ATE) as the evaluation metrics, which includes RMSE Trans (m) and RMSE Rot (deg). Table \ref{tab: res_kaist} displays the results of the comparison of our algorithm to the baselines in various scenarios, indicating that our proposed algorithm achieved comparable localization accuracy to the baselines.

\begin{table}[htbp]
    \caption{RMSE results of KAIST urban dataset.}
    \label{tab: res_kaist}
    \begin{center}
        \begin{tabular}{cccc}
            \bottomrule
            \makecell[c]{dataset                                              \\ Trans (m)\\ Rot (deg)} & Suburban    & Urban & Highway   \\ \hline
            CL+PA \cite{liao2020coarse}         & \makecell[c]{0.604          \\ 0.882} & \makecell[c]{0.580\\ 1.080} & \makecell[c]{1.806\\ 0.935} \\ \hline
            PC Semantic \cite{stenborg2018long} & \makecell[c]{1.798          \\ 0.464} & \makecell[c]{0.893\\ 0.91}  & \makecell[c]{2.494\\ 0.907} \\ \hline
            fusion SFM \cite{li2020vision}      & \makecell[c]{0.573          \\ 0.510} & \makecell[c]{0.54\\ 0.68}   & \makecell[c]{1.964\\ 0.853} \\ \hline
            Ours                                & \makecell[c]{\textbf{0.525} \\ \textbf{0.507}} & \makecell[c]{\textbf{0.472}\\ \textbf{0.701}} & \makecell[c]{\textbf{1.673}\\ \textbf{0.822}} \\ \bottomrule
        \end{tabular}
    \end{center}
\end{table}

To further evaluate the effectiveness and generalizability of our system, we conduct an experiment base on our self-recorded dataset and compared our algorithm to the state-of-the-art visual localization toolbox Hloc \cite{sarlin2019coarse}. 
We follow the standard evaluation method proposed in \cite{sattler2018benchmarking} for outdoor localization: $(0.25m, 2^{\circ})$, $(0.5m, 5^{\circ})$ and $(5m, 10^{\circ})$. Detailed results of the comparison with Hloc are presented in Table \ref{tab: res_park}. Notably, for the park dataset, Hloc requires an additional storage of about 4.5G of map data in colmap \cite{schoenberger2016sfm} format, while our system only needs to keep about 2M semantic point cloud maps. Despite the much smaller prior map, our proposed system achieves higher translation and rotation accuracy than the baseline.
In addition, we observe that the overall localization accuracy in the industrial park are not as good as the public road due to the incomplete and scarce lane markings, as shown in Fig. \ref{fig:illus_loc}(b). In contrast, Hloc can achieve higher accuracy than vacant public roads with the help of features such as dense buildings. Fig. \ref{fig:illus_loc}(c) and (d) illustrate visual examples of our localization algorithm running in real time based on the park and the public road dataset.

\begin{table}[htbp]
    \caption{Performance comparison of proposed algorithm for self-recorded dataset.}
    \label{tab: res_park}
    \begin{center}
        \begin{tabular}{ccccc}
            \bottomrule
            dataset                     & \multicolumn{2}{c}{Park} & \multicolumn{2}{c}{Public Road} \\ \cline{2-5}
            \makecell[c]{m                                                                           \\ deg}   & \makecell[c]{0.25/0.5/5.0\\ 2/5/10}   & \makecell[c]{Trans\\ Rot} & \makecell[c]{0.25/0.5/5.0\\ 2/5/10}   & \makecell[c]{Trans\\ Rot} \\ \hline
            Hloc\cite{sarlin2019coarse} & 30.47/63.49/95.5l        & \makecell[c]{1.25               \\ 0.59} & 26.33/56.32/93.58 & \makecell[c]{1.43\\ 0.71} \\ \hline
            Ours                        & 38.38/77.23/97.72        & \makecell[c]{0.52               \\ 0.63} & 32.86/80.16/98.21 & \makecell[c]{0.49\\ 0.65} \\ \bottomrule
        \end{tabular}
    \end{center}
\end{table}

\begin{figure}[htbp]
    \centering
    \includegraphics[width=0.45\textwidth]{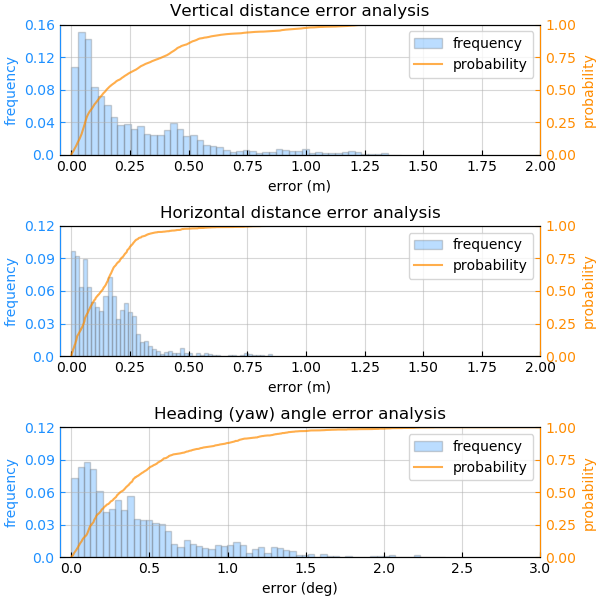}
    \caption{The probability distribution plot of localization error in vertical and horizontal direction, and heading angle respectively.}
    \label{fig:err_anal}
\end{figure}

Fig. \ref{fig:err_anal} illustrates the distribution of vertical and horizontal position error in the vehicle frame and the heading angle error of our system. In comparison, the horizontal error distribution is more concentrated and closer to zero, which confirms that lane markings particularly the prevailing lane line feature, have stronger constraints on the horizontal direction. Poor accuracy in the vertical direction and heading angle may result from a lack of pole supervision in certain cases.

To evaluate the effectiveness of each proposed features in detail, we conduct an ablation study on our public road dataset. To ensure a fair comparison with VIO results, we also consider the relative pose error (RPE) metric using EVO \cite{grupp2017evo}, as shown in Table \ref{tab: res_ablation}. Our method eliminates the cumulative drift error of VIO by incorporating the global map, leading to a translational RMSE of 0.492 m, which is acceptable for autonomous driving tasks. Interestingly, both the lane markings and pole features of the semantic map outperform the baseline in terms of RPE, indicating that visual features contribute to more efficient and robust localization accuracy.

\begin{table}[htbp]
    \caption{Validation results of different methods on public road dataset.}
    \label{tab: res_ablation}
    \begin{center}
        \begin{tabular}{c|cc|cc}
            \hline
            VIO        & lane markers & poles      & \makecell[c]{ATE Trans         \\ m}   & \makecell[c]{RPE Trans\\ m}   \\ \hline
            \checkmark &              &            & 152.52                 & 0.096 \\
            \checkmark & \checkmark   &            & 0.513                  & 0.041 \\
            \checkmark &              & \checkmark & 0.546                  & 0.043 \\
            \checkmark & \checkmark   & \checkmark & 0.492                  & 0.038 \\ \hline
        \end{tabular}
    \end{center}
    "\checkmark" means the corresponding feature is selected.
\end{table}

\section{CONCLUSIONS}
In this paper, we propose a visual localization system for autonomous vehicles based on stable visual semantic features, such as ground markers, lane lines, and poles. In our framework, we first construct the semantic map offline using LiDAR, and then optimize the matching of semantic features and corresponding information from the map to estimate the current position and direction of the vehicle. We validate our proposed system in a variety of challenging real-world traffic scenarios, and the results show that our proposed approach achieves better translation and rotation accuracy than the baseline. In future work, we consider integrating more kinds of low-cost sensors, such as GPS, to further extend the application of robust localization of autonomous vehicles in more complex traffic scenarios.










\bibliographystyle{ieeetr}
\bibliography{bib}

\end{document}